\documentclass[a4paper, 10pt, conference]{ieeeconf}      
\usepackage{cite} 
\usepackage{graphics} 
\usepackage{epsfig} 
\usepackage{mathptmx} 
\usepackage{times} 
\usepackage{amsmath} 
\usepackage{amssymb}  
\usepackage{longtable}
\usepackage{lscape}
\usepackage{multirow}
\usepackage{subfigure}
\usepackage{array}
\usepackage{url}
\usepackage{lipsum}
\usepackage{mathtools}
\usepackage{cuted}
\usepackage[utf8]{inputenc}
\usepackage{FG2019}

\FGfinalcopy 

\IEEEoverridecommandlockouts                              
\def\FGPaperID{****} 

\title{\LARGE \bf
Spotting Micro-Expressions on Long Videos Sequences
}

\author{\parbox{16cm}{\centering
    {\large Jingting Li$^1$, Catherine Soladié $^1$, Renaud Séguier $^1$, Su-Jing Wang$^2$ and Moi Hoon Yap$^3$}\\
    {\normalsize
    $^1$ CentraleSupélec, CNRS, IETR, UMR 6164, F-35000 Rennes, France\\
    $^2$ Key Laboratory of Behavior Sciences, Institute of Psychology, Chinese Academy of Sciences, Beijing, 100101, China\\
    $^3$ Manchester Metropolitan University, Manchester, M1 5GD, UK}}
    \thanks{This work is supported by Chinese scholarship council and ANR reflet. This paper is also supported in part by grants from the National Natural Science Foundation of China (61772511) and The Royal Society (IF160006).}
}

\begin{document}
\IEEEoverridecommandlockouts\pubid{\makebox[\columnwidth]{978-1-7281-0089-0/19/\$31.00~\copyright{}2019 IEEE \hfill}
\hspace{\columnsep}\makebox[\columnwidth]{ }}

\ifFGfinal
\thispagestyle{empty}
\pagestyle{empty}
\else
\author{Anonymous FG 2019 submission\\ Paper ID \FGPaperID \\}
\pagestyle{plain}
\fi
\maketitle

\begin{abstract}

This paper presents two methods for the first Micro-Expression Spotting Challenge 2019 by evaluating local temporal pattern (LTP) and local binary pattern (LBP) on two most recent databases, i.e. SAMM and CAS(ME)$^2$. First we propose LTP-ML method as the baseline results for the challenge and then we compare the results with the LBP-$\chi ^{2}$-distance method. The LTP patterns are extracted by applying PCA in a temporal window  on several facial local regions. The micro-expression sequences are then spotted by a local classification of LTP and a global fusion. The LBP-$\chi ^{2}$-distance method is to compare the feature difference by calculating $\chi^2$ distance of LBP in a time window, the facial movements are then detected with a threshold.
The performance is evaluated by Leave-One-Subject-Out cross validation.
The overlap frames are used to determine the \textit{True Positives} and the metric \textit{F1-score} is used to compare the spotting performance of the databases. The \textit{F1-score} of LTP-ML result for SAMM and CAS(ME)$^2$ are 0.0316 and 0.0179, respectively. The results show our proposed LTP-ML method outperformed LBP-$\chi ^{2}$-distance method in terms of \textit{F1-score} on both databases. 

\end{abstract}

\section{INTRODUCTION}
Facial micro-expression (ME) is a local brief facial movement, which can be triggered under high emotional pressure. The duration is less than 500ms~\cite{Ekman_Friesen_1969}. It is a very important non-verbal communication clue, the involuntary nature make it possible to analyze personal genuine emotional state. ME analysis has many potential applications in national security~\cite{Ekman_2009}, medical care~\cite{Endres_Laidlaw_2009}, educational psychology~\cite{Chiu_Liaw_Yu_Chou}, and political psychology~\cite{Stewart_Waller_Schubert_2009}. Due to the growth and importance of MEs, researchers~\cite{yap2018facial} have worked collaboratively to solicit the works in this area by conducting challenges in datasets and methods for MEs. This year, the theme of the Second Facial Micro-Expression Grand Challenge has extended to spotting challenges.
\par
The main idea of most methods for ME spotting is to compare the feature differences between the first frame and the other frames in a time window. 
Meanwhile, the feature descriptors used in the state of the art are diverse, to name a few: LBP~\cite{Moilanen_Zhao_Pietikainen_2014,Li_Hong_Moilanen_Huang_Pfister_Zhao_Pietikainen_2017}, HOG~\cite{davison2018objective}, optical flow~\cite{yan2014casme, Li_Yu_Zhan_2016,Wang_Wu_Fu_2016,ma2017region}, integral projection~\cite{lu2017micro}, Riesz pyramid~\cite{duque2018micro}, and frequency domain~\cite{li2018can}. Feature differences allow consistent comparisons between frames over a time window of the size of an ME. However, the movements spotted between frames might not be the ME movements, it could be noises, macro-movements and illumination changes. This is why the ability to distinguish MEs from other movements (such as blinking or subtle head movements) remains an open challenge.
\par
Nowadays, methods utilizing machine learning are emerging~\cite{Xia_Feng_Peng_Peng_Zhao_2016,tran2017sliding,borza2017high,husak2017spotting}. Furthermore,~\cite{zhang2018smeconvnet} employed deep learning for the first time to perform the ME spotting.
The machine learning process enhances the ability of distinguishing micro-expression from others. However, the spatial patterns are still the primary feature for the classifier. The temporal variation pattern of facial movement in a ME duration has yet to attract sufficient attention. Meanwhile, few articles spotted micro-expression directly from local region. However, the characteristic of that the micro-expression is a local facial movement could help to reduce the false positives. 
\par
In this paper, we spot the micro-expression clips in two recently published databases, and establish the baseline method for ME spotting challenge by using directly a temporal pattern extracted from local region~\cite{li2018ltp}. Frames in a ME duration are taken into account to obtain a real temporal and local pattern (LTP), and then the LTPs are recognized by a classifier. Even though the spatial pattern is not studied, the spotted facial motions are differentiated by a fusion process from local to global. This method helps to improve the ability to distinguish ME from other movements. Furthermore, it allows finding the ME spatial local region and the temporal onset index of ME. We compare the results of our proposed LTP-ML method with a LBP approach - LBP-$\chi ^{2}$-distance by Moilanen et al.~\cite{Moilanen_Zhao_Pietikainen_2014}.
\par
The rest of the paper is organized as follows:
Section~\ref{sec:method} presents the methodology and performance metrics. Section~\ref{sec:result} introduces the result and also shows the detailed experiment results. Section~\ref{sec:conclusion} concludes the paper.

\section{Methodology}
\label{sec:method}
This section describes the benchmark databases, the proposed LTP-ML method, the state-of-the-art LBP method and the performance metrics.
\subsection{Databases}
Two most recent long videos spontaneous micro-expression databases, SAMM~\cite{davison2018samm} and CAS(ME)$^2$~\cite{qu2017cas}, are used for ME spotting challenge. Both databases contain long videos, which were recorded in the strictly controlled laboratory environment. Table~\ref{tab:bdd_info} compares the differences between these two databases. The notable differences are the resolution and frame rates used in the experimental settings. These are indeed a great challenge for computer vision and machine learning community to produce a robust method worked for both databases, The detailed information of these two databases is presented in the following two subsections. 

\begin{table}[h]
 \caption{A Comparison between SAMM and CAS(ME)$^2$.}
    \centering
    \begin{tabular}{|c|c|c|c|c|}
    \hline
        Database & Participants&Samples&Resolution &FPS  \\
        \hline
        SAMM &32 &79&2040$\times$1088&200\\
        \hline
        CAS(ME)$^2$   &22&97&640$\times$480&30 \\
        \hline
    \end{tabular}
   
    \label{tab:bdd_info}
\end{table}
\begin{figure}[thpb]
\subfigure{
    \begin{minipage}{0.22\textwidth}
    \includegraphics[width = \textwidth,height = 4cm]{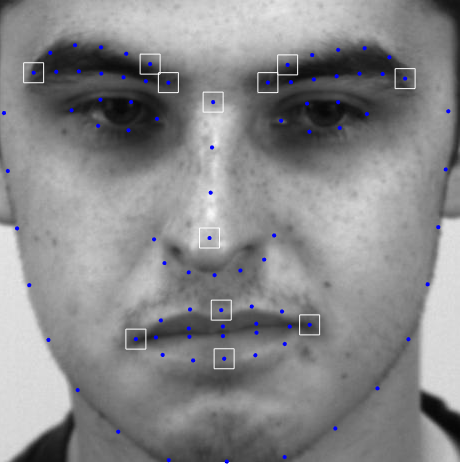}
    \end{minipage}
}
    \subfigure{
       \begin{minipage}{0.22\textwidth}
    \includegraphics[width = \textwidth,height = 4cm]{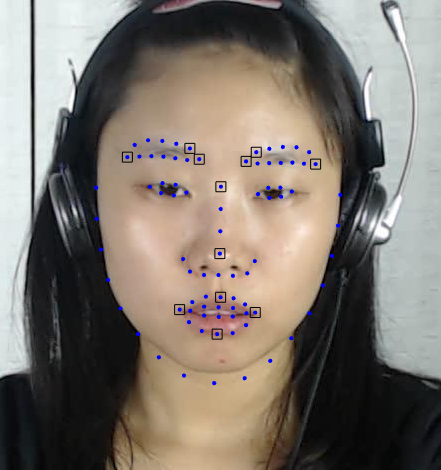}
        \end{minipage}
    }
    \caption{Facial landmarks tracking and ROI selection. On the left: an example from SAMM; on the right: an example from CAS(ME)$^2$ }
    \label{fig:landmarks_track}
\end{figure}

\subsubsection{SAMM Long Videos Database}
SAMM database consists of a total of 32 subjects and each has 7 videos~\cite{davison2018samm}. The average length of videos is 35.3s. The original release of SAMM consists of micro-movement clips labelled in Action Units. Recently, the authors \cite{Davison2018} introduced objective classes and emotion classes for the database. The recognition challenge will be using the emotional classes from the database as ground truth. The spotting challenge focuses on 79 videos, each contains one/multiple micro-movements, with a total of 159 micro-movements. The index of onset, apex and offset frames of micro-movements were provided as the ground truth. The micro-movements interval is from onset frame to offset frame. In this database, all the micro-movements are labeled. Thus, the spotted frames can indicate not only ME but also other facial movements, such as eye blinks.

\subsubsection{CAS(ME)$^2$ Database}
In the part A of CAS(ME)$^2$ database~\cite{qu2017cas}, there are 22 subjects and 97 long videos. The average duration is 148s. The facial movements are classified as macro- and micro-expressions. The video samples may contain multiple macro or micro facial expressions. The onset, apex, offset index for these expressions are given in the excel file. In addition, the eye blinks are labeled with onset and offset time.

\subsection{LTP-ML: Our Proposed Baseline Method}
The baseline method is developed based on the proposed LTP-ML (local temporal pattern-machine learning) method in ~\cite{li2018ltp}. The method is extended for long videos by employing a sliding temporal window. The main idea and the modification of LTP-ML method is presented in the following paragraphs.    
\subsubsection{Pre-processing}

As the ME is a local facial movement, we analyze ME only on a selection of regions of interest (ROIs). First of all, as shown in Figure~\ref{fig:landmarks_track}, 84 facial landmarks are tracked in the video sequence by utilizing the Genfacetracker (\copyright Dynamixyz). Then the size of ROI square $ a $ is determined by the distance $ L$ between the left and right inner corners of eyes: $ a = (1/5) \times L$. 12 ROIs squares are chosen based on the regions where ME happens most frequently, i.e. the corner of the eyebrows and of the mouth. Two ROIs of nose region are chosen as references because the nose is the most rigid facial region. 
\par

\begin{figure}[thpb]
    \centering
    \includegraphics[width=.5\textwidth]{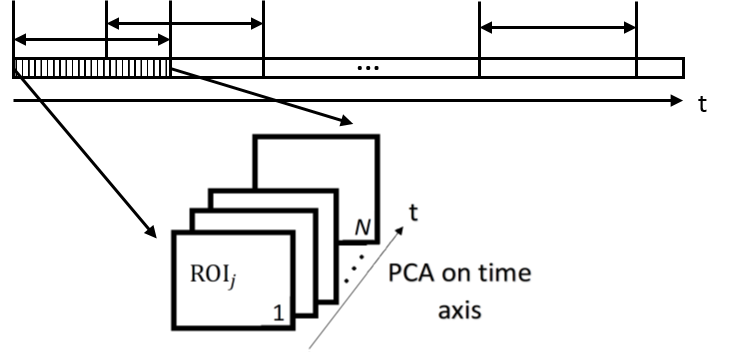}
    \caption{PCA process analysis. The long video is divided into small sequences by a sliding window. Then the PCA process is performed respectively on time axis for 12 ROIs sequences in one small divided clip. }
    \label{fig:PCA}
\end{figure}

Since the average duration of ME is around 300ms, and the subjects barely moved in one second, the long videos in these two databases are processed by a temporal sliding window $W_{video}$ whose length is 1s. The overlap is set to 300ms to avoid missing any possible ME movements. This, the video is separated into an ensemble of small sequences $[I_1, I_2,...,I_M] $ by sliding temporal window as shown in Figure~\ref{fig:PCA}. The positions of 12 chosen ROIs for all frames in one sequence are determined by the detected landmarks of the first frame in the window.

\subsubsection{Feature Extraction}
In this part, local temporal patterns (LTPs)~\cite{li2018ltp} are analyzed in the local region to distinguish ME from other movements. They are extracted from 12 ROIs respectively in each small sequence. Supposing in sequence $I_m$ ($m \leq M$), as illustrated in the lower part of Figure~\ref{fig:PCA}, PCA is performed on the temporal axis of each ROI sequence to conserve the principal variation at this region. The first two components of each ROI frame are used to analyze the variation pattern of local movement. The PCA process for ROI sequence $ROI^m_j$ ($j \leq 12 $) in $I_m$ can be presented as in equation \ref{eq:one}. 

\begin{strip}
\begin{equation}
\label{eq:one}
\resizebox{0.7\textwidth}{!}{$
  \left[
\begin{matrix}
   P_{1}^{m,j}(x) &  \cdots & P_{N}^{m,j}(x)\\
   P_{1}^{m,j}(y) & \cdots & P_{N}^{m,j}(y)   
  \end{matrix}
  \right]= \Phi \times ( 
  \left[
  \begin{matrix}
   F_{1}^{m,j}(1) & \cdots & F_{N}^{m,j}(1)\\
     & \ddots &    \\
    F_{1}^{m,j}(a^{2}) & \cdots & F_{N}^{m,j}(a^{2})   
  \end{matrix}
   \right]-\bar I )  $}
\end{equation}
\end{strip}

where $F_{n}^{m,j}$ represents the pixels in one ROI frame, $P_{n}^{m,j} = [P_{n}^{m,j}(x),P_{n}^{m,j}(y)]$ are the first two components of PCA, $n$ is the frame index in this ROI sequence ($n \leq N $). Hence, each frame in $ROI^m_j$ can be represented by a point $P_{n}^{m,j}$. \par

\begin{figure}[thpb]
    \centering
    \includegraphics[width = 0.5\textwidth]{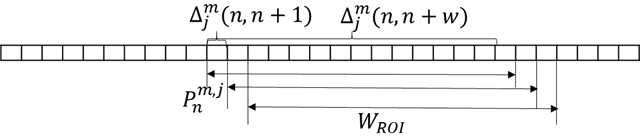}
    \caption{Distance calculation for one ROI sequence $ROI^m_j$ in video clip $I_m$. }
    \label{fig:dist_window}
\end{figure}

Then, a sliding window $W_{ROI}$ is set depending on the average duration of ME (300ms). The distances between the first frame and the other frames in this window are calculated. The window goes through each frame in the sequence  $ROI^m_j$, and the distance set can be got as $[\Delta^m_j(n,n+1),\Delta^m_j(n,n+w),...,\Delta^m_j(n,n+W_{ROI}-1) ] $, as shown in Figure~\ref{fig:dist_window}.
\par
The values of distance are then normalized for the entire $ROI^m_j$ to avoid the influence of different movement magnitude in different videos. 
Hence, the feature of frame n for $ROI^m_j$ can be represented as: $[CN^m_j$, $ d^m_j(n,n+1),\cdots,d^m_j(n,n+W_{ROI}-1) ]$, where $d^m_j(n,n+1)$ is the normalized distance value and the $CN^m_j$ is the normalization coefficient. The more detailed deduction process can be found in~\cite{li2018ltp}. The feature for one ROI sequence of the entire long video is the concatenation of features of all the separated sequences. 

\subsubsection{Local Classification}
As presented in the above paragraph, one video contains 12 feature ensembles from 12 ROI. Li et al.~\cite{li2018ltp} showed the LTP patterns are similar for all chosen ROIs for all kinds of ME. The patterns which can represent the ME local movements can be recognized by a local classification. A supervised classification SVM is employed with Leave-One-Subject-Out cross validation. The feature selection and label annotation are presented in~\cite{li2018ltp}.  

\subsubsection{Global Fusion}
After the LTPs which fit the local ME movement pattern are recognized, a global fusion is processed to eliminate the false positives concerning other movements and true negatives caused by our recognition process. As introduced in~\cite{li2018ltp}, there are three steps: a local qualification, a spatial fusion and a merge process. 

\subsection{LBP-$\chi ^{2}$-distance Method}
This method is firstly proposed in ~\cite{Moilanen_Zhao_Pietikainen_2014}. It is the most commonly used method for result comparison for ME spotting.
Based on~\cite{Moilanen_Zhao_Pietikainen_2014} and~\cite{tran2017sliding}, the configuration of LBP-$\chi^2$ is set as follows: the entire face region is divided into 36 blocks. The overlap rates between blocks on axis X and Y are are 0.2 and 0.3 respectively. LBP features are extracted from blocks with uniform mapping. The radius r is set to $ r = 3 $, and the number of neighboring points p is set to $ p = 8 $. The $ \chi^{2} $  distances of the each frame are computed in an $ 2 \times L_{interval}+1 $ interval.
\par
First of all, the value of LBP-$\chi^2$-distance is compared in the whole long video. However, the method can barely spot any micro-expression intervals, while there are many false positives. This is due to this method spots the maximal movements in the video, and there are some larger movements than ME in both databases. Hence, the entire video is separated into a sub-video set by a sliding window, the setting is the same as the LTP-ML method. For each sub-video, the feature differences are calculated and sorted to find the maximal movement in this short interval. This gives the chance to spot more MEs which could be ignored in entire video comparison. 


\begin{table*}[h]
\caption{Baseline result for micro-expression spotting. SAMM$_{ME}^{c}$ represents the SAMM cropped-face videos contain ME, SAMM$_{ME}^{f}$ are the ME videos with full frame, CAS(ME)$^2_{ME}$ means all the videos in this sub-dataset of CAS(ME)$^2$ have ME sequences. }
    \centering
    \begin{tabular}{|c|c|c|c|c|c|c|c|c|}
    \hline
        Method & \multicolumn{4}{|c|}{LTP-ML}& \multicolumn{3}{|c|}{LBP-$\chi^2$}\\
        \hline 
        database&SAMM$_{ME}^{c}$ &SAMM$_{ME}^{f}$ &CAS(ME)$^2_{ME}$ &CAS(ME)$^{2}$& SAMM$_{ME}^{c}$ &CAS(ME)$^2_{ME}$ &CAS(ME)$^{2}$ \\
        \hline
        nb\_vid&79& 79&32&97 &79&32&97\\
        \hline
         TP&34& 47&16&16&12 &10 &10 \\
         \hline
        FP&1958&3891 &1711&5742&4172 &1729 & 5435\\
        \hline
        FN& 125& 112&41&41 &147 &47 &47 \\
        \hline
        Precision&0.0171&0.0043 &0.0093&0.0028&0.0028 & 0.0057&0.0018 \\
        \hline
         Recall&0.2138& 0.2956&0.2807&0.2807&0.0755 &0.1754 & 0.1754\\
         \hline
         F1-score&\textbf{0.0316}&0.0229 &\textbf{0.0179}&\textbf{0.0055}& 0.0055& 0.0111& 0.0035\\
       
         \hline
    \end{tabular}
    
    \label{tab:result}
\end{table*}

\subsection{Performance Metrics}
There are three evaluation methods used to compare the performance of the spotting tasks: \par
\textbf{1. True positive in one video definition}
Supposing there are $m$ micro-expressions in the video, and $n$ intervals are spotted. 
The result of this spotted interval $W_{spotted}$ is considered as \textit{true positive (TP)} if it fits the following condition:
\begin{equation}
   \frac{W_{spotted}\cap W_{groundTruth}}{W_{spotted}\cup W_{groundTruth}} \geq k
\end{equation}
where $k$ is set to 0.5, $W_{groundTruth}$ represents the micro-expression interval (onset-offset).
Otherwise, the spotted interval is regarded as \textit{false positive (FP)}.
\par
\textbf{2. Result evaluation in one video}
Supposing the number of \textit{TP} in one video is $a$ ($a\leq m$ and $a\leq n$), then \textit{FP} = $n-a$, \textit{false positive (FN)} = $m-a$, the \textit{Recall}, \textit{Precision} and \textit{F1-score} are defined:
\begin{equation}
    Recall = \frac{a}{m},\  Precision = \frac{a}{n}
\end{equation}

\begin{equation}
    F-score = \frac{2TP}{2TP+FP+FN} = \frac{2a}{m+n}
\end{equation}
\par
In practical, these metrics might not be suitable for some videos, as there exist the following situations on a single video:
\begin{itemize}
    \item The test video does not have micro-expression sequences, thus, $m=0$, the denominator of recall will be zeros. 
    \item The spotting method does not spot any intervals. The denominator of precision will be zeros since $n=0$.
    \item If there are two spotting methods, Method$_1$ spots p intervals and Method$_2$ spots q intervals, and $p\leq q$. Supposing for both methods, the number of true positive is 0, thus the metrics (\textit{recall}, \textit{precision} or \textit{F1-score}) values both equal to zeros. However, in fact, the Method$_1$ performs better than Method$_2$. 
\end{itemize}
\par
Considering these situations, we propose for a single video, we record the result in terms of \textit{TP}, \textit{FP} and \textit{FN}. For performance comparison, we produce a final calculation of other metrics for the entire database.
\par
\textbf{3. Evaluation for entire database}
Supposing in the entire database, there are $V$ videos and $M$ micro-expression sequences, and the method spot $N $ intervals in total. The database could be considered as one long video, thus, the metrics for entire database can be calculated by:
\begin{equation}
Recall_{D} = \frac{{{\sum^V_{i=1}}} a_i}{{{\sum^V_{i=1}}} m_i} = \frac{A}{M}
\end{equation}
\begin{equation}
Precision_{D} = \frac{{{\sum^V_{i=1}}} a_i}{{{\sum^V_{i=1}}} n_i} = \frac{A}{N}
\end{equation}
\begin{equation}
F1-score_{D} = \frac{2\times (Recall_{D} \times Precision_{D})} {Recall_{D}+Precision_{D}} 
\end{equation}
\par
The final results by different methods would be evaluated by \textit{F1-score} since it considers the both \textit{recall} and \textit{precision}. 

\section{Results and Discussion}
\label{sec:result}
As introduced in Section~\ref{sec:method}, SAMM and CAS(ME)$^2$ have different frame rates and resolution. Hence, the lengths of sliding window $W_{video}$, the overlap size, the interval length of $W_{ROI}$ and the ROIs size are different for these two databases. Table~\ref{tab:exp_conf} lists the experimental parameters.
\par
\begin{table}[h]
\caption{Parameter configuration for SAMM and CAS(ME)$^2$. $L_{window}$ is the length of sliding window $W_{video}$,$L_{overlap}$ is the overlap size between sliding windows, $L_{interval}$ is the interval length of $W_{ROI}$. }
    \centering
    \begin{tabular}{|c|c|c|c|c|}
    \hline
        Database & $L_{window}$& $L_{overlap}$& $L_{interval}$ &$size_{ROI}$ \\
        \hline
         SAMM & 200&60 &60&15\\
         \hline
          CAS(ME)$^2$&30&9& 9&10\\
          \hline
    \end{tabular}
    \label{tab:exp_conf}
\end{table}

For CAS(ME)$^2$ database, there are 97 videos, but only 32 videos contain micro-expressions. Thus, different results are given under two conditions: one is only considering 32 videos which have ME (CAS(ME)$^2_{ME}$), another one is to include the entire database (all 97 videos). Since the raw videos in SAMM database are too big to download (700GB), only 79 videos (full frame: 270GB and cropped face: 11GB) were provided for the challenge. In this work, we report the results based on these two versions of SAMM database: one is the cropped videos (SAMM$_{ME}^{c}$) provided by the authors using the method in~\cite{Davison_Yap_Lansley_2015}, and the other one is the videos with full frame (SAMM$_{ME}^{f}$). The spotting process is performed only on the downloaded databases.  
\par
\subsection{Experiments Results of LTP-ML Method}
After performing the LTP-ML method on these two databases, the spotting results for whole database are listed in Table~\ref{tab:result}.
The \textit{F1-score} for (SAMM$_{ME}^{c}$) and CAS(ME)$^2_{ME}$ are 0.0316 and 0.0179 respectively. LTP-ML performs better in SAMM$_{ME}^{c}$ than SAMM$_{ME}^{f}$, since the cropped-face process has already aligned the face region in the video, and reduced the influence of irrelevant movements. Concerning the spotting result of CAS(ME)$^2$, there are more \textit{FPs} because the video in this database which has no ME may contain macro-expressions. 

\par

\subsection{Experiments Results of LBP-$\chi^2$-distance (LBP-$\chi^2$) Method}
The result is compared with LBP-$\chi^2$-distance (LBP-$\chi^2$) method. The spotting result is listed in Table~\ref{tab:result}. For CAS(ME)$^2_{ME}$, when the threshold for peak selection is set to 0.15, we can get the best result for LBP-$\chi^2$ method, the \textit{F1-score} is 0.0111. Meanwhile, the highest \textit{F1-score} of SAMM$_{ME}^{c}$ is 0.0055 when the threshold is set to 0.05. 
\par
Compared with LTP-ML method, LBP-$\chi^2$ method is less accurate. LTP-ML method is capable of spotting the subtle movements based on the patterns which represented the temporal pattern variation of ME. Yet, the value of \textit{F1-score} is low because of the large amounts of \textit{FP}. Both databases contain noises and irrelevant facial movements, especially for CAS(ME)$^2$, it is not easy to separate macro-expressions from micro-expressions based on 30fps videos. The ability of distinguishing ME from other movements still need to be enhanced. 

\section{CONCLUSIONS}
\label{sec:conclusion}
This paper addresses the challenge in spotting ME on long videos sequences using two most recent databases, i.e. SAMM and CAS(ME)$^2$. We proposed LTP-ML for spotting MEs and provided a set of performance metrics as the guideline for result evaluation on ME spotting. The baseline results of these two databases are provided in this paper. We demonstrate that our proposed method is better than the LBP approach in spotting MEs. Whilst the method was able to produce a reasonable amount of \textit{TPs}, there are still a huge challenge lays ahead due to the large amount of \textit{FPs}. Further research will focus on enhancing the ability of distinguishing ME from other facial movements to reduce \textit{FPs}, including the implementation of deep learning approaches when we have sufficient data.

\section{ACKNOWLEDGMENTS}

The authors gratefully acknowledge the contribution of the Organisers and Program Committee Members.


\bibliographystyle{IEEEtran}
\bibliography{IEEEabrv,reference}

\end{document}